\RequirePackage{fix-cm}
\documentclass[twocolumn]{svjour3}          
\smartqed  
\usepackage{graphicx}
\usepackage{graphicx}
\usepackage{multirow}
\usepackage{amsmath}
\usepackage{amssymb}
\usepackage{bm}
\usepackage{bbm}
\usepackage{comment}
\usepackage{algorithm}
\usepackage{listings}
\usepackage{subfig}
\usepackage{booktabs}
\usepackage{array} 
\usepackage[pagebackref=false,breaklinks=true,colorlinks,bookmarks=false]{hyperref}
\usepackage[misc]{ifsym}

\usepackage[dvipsnames]{xcolor}
\newcommand{\blue}[1]{#1}
\newcommand{\green}[1]{#1}
\newcommand{\red}[1]{#1}
\newcommand{\orange}[1]{#1}

\newcommand{\romannum}[1]{\romannumeral #1}
\newcommand{\minisec}[1]{\vspace{0.1cm}\noindent\textbf{#1}\quad}

\newcommand{\tableCellHeight}{1}
\newcommand{\tabstyle}[1]{
  \setlength{\tabcolsep}{#1}
  \renewcommand{\arraystretch}{\tableCellHeight}
  \centering
}

\begin{document}
\sloppy

\title{MixStyle Neural Networks for Domain Generalization and Adaptation
}


\author{Kaiyang Zhou \and
        Yongxin Yang \and
        Yu Qiao \and
        Tao Xiang
}


\institute{Kaiyang Zhou (\Letter) \at
           Hong Kong Baptist University, Hong Kong SAR, China \\
           \email{kyzhou@hkbu.edu.hk}
           \and
           Yongxin Yang \at
           Queen Mary University of London, UK \\
           \email{yongxin.yang@qmul.ac.uk}
           \and
           Yu Qiao \at
           Shanghai AI Laboratory \& Shenzhen Institute of Advanced Technology, CAS \\
           \email{yu.qiao@siat.ac.cn}
           \and
           Tao Xiang \at
           University of Surrey, UK \\
           \email{t.xiang@surrey.ac.uk}
}

\date{Received: date / Accepted: date}

\maketitle

\begin{abstract}
Neural networks do not generalize well to unseen data with domain shifts---a longstanding problem in machine learning and AI. To overcome the problem, we propose MixStyle, a simple plug-and-play, parameter-free module that can improve domain generalization performance without the need to collect more data or increase model capacity. The design of MixStyle is simple: it mixes the feature statistics of two random instances in a single forward pass during training. The idea is grounded by the finding from recent style transfer research that feature statistics capture image style information, which essentially defines visual domains. Therefore, mixing feature statistics can be seen as an efficient way to synthesize new domains in the feature space, thus achieving data augmentation. MixStyle is easy to implement with a few lines of code, does not require modification to training objectives, and can fit a variety of learning paradigms including supervised domain generalization, semi-supervised domain generalization, and unsupervised domain adaptation. Our experiments show that MixStyle can significantly boost out-of-distribution generalization performance across a wide range of tasks including image recognition, instance retrieval and reinforcement learning. The source code is released at \url{https://github.com/KaiyangZhou/mixstyle-release}.
\end{abstract}

\section{Introduction}\label{sec:introduction}

Convolutional neural networks (CNNs) have been a key driving force behind successes in computer vision over the past decade~\cite{krizhevsky2012imagenet,he2016deep,szegedy2015going}. For example, given millions of labeled images for training, CNNs can surpass humans in classifying images into 1,000 categories on ImageNet~\cite{he2016deep}. In reinforcement learning (RL), CNNs have also shown remarkable ability in extracting meaningful representations enabling RL agents to play Atari games at near-human level~\cite{mnih2013playing}. However, these achievements are largely limited to independent and identically distributed (i.i.d.) datasets. When it comes to out-of-distribution (OOD) target datasets, which are common in practice, CNNs often fail catastrophically with low performance that is far away from that seen in source datasets~\cite{zhou2021domain}. This significantly hinders their deployment in real-world applications.

To mitigate domain discrepancy between training and test data, a straightforward solution is to collect diverse source data from multiple relevant domains so a CNN model is allowed to intrinsically learn more domain-invariant, and hence generalizable representations. This has long been studied under domain generalization (DG)~\cite{zhou2021domain,blanchard2011generalizing,muandet2013domain,fan2022normalization}. However, DG models in practice are often constrained by the amount (and diversity) of source domains. The reason is clear: collecting more training data with annotation is expensive and time-consuming; this is further compounded when the data has to come from multiple and diverse domains. Recent studies have shown promising results with neural networks-based data augmentation methods, such as perturbing images with adversarial gradients~\cite{shankar2018generalizing} or learning neural networks to generate novel-looking images~\cite{zhou2020learning}. Such a model-based data augmentation~\cite{zhou2021domain} approach is, however, less appealing to real-world deployment because it often consumes considerable compute resources in terms of GPU memories and training hours, as well as considerable engineering efforts.

\begin{figure}[t]
    \centering
    \includegraphics[width=\columnwidth]{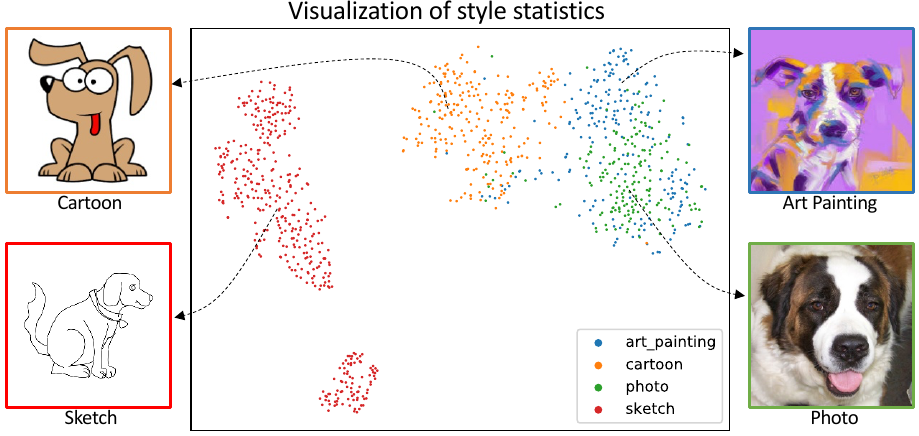}
    \caption{2D t-SNE~\cite{tsne} visualization of style statistics (concatenation of means and standard deviations) computed from the first residual block's feature maps of a ResNet-18 ~\cite{he2016deep} trained on four distinct domains~\cite{li2017deeper}. The idea of our MixStyle approach is to mix these instance-level feature statistics to efficiently synthesize novel domains.}
    \label{fig:motivation}
\end{figure}

In this work, we propose \emph{MixStyle}, a plug-and-play, parameter-free module that can be simply inserted to shallow CNN layers to achieve implicit data augmentation and requires no modification to training objectives. Specifically, MixStyle probabilistically mixes the feature statistics---i.e., means and standard deviations of feature maps---between random instances. The idea is inspired by the observation that visual domains can often be characterized by image styles, which are in turn encapsulated within instance-level feature statistics in shallow CNN layers~\cite{huang2017arbitrary,dumoulin2017learned}. An example is shown in Fig.~\ref{fig:motivation}: the four images depict the same semantic concept, i.e.~dog, but have distinctive styles (e.g., characteristics in colors and textures); and the feature statistics clearly capture these styles reflected in the separable clusters. Therefore, MixStyle essentially synthesizes novel domains. Though designed for tackling the DG problem, MixStyle can also be applied for problems where unlabeled images are available, e.g., semi-supervised domain generalization~\cite{zhou2021stylematch} and unsupervised domain adaptation~\cite{ganin2015unsupervised}. This is achieved by introducing a simple extension by mixing feature statistics between labeled and pseudo-labeled instances. It is worth noting that MixStyle \romannum{1}) is architecture-agnostic, \romannum{2}) does not necessarily require domain labels as combining instance styles can also lead to new styles, \romannum{3}) perfectly fits into mini-batch training, and \romannum{4}) is easy to implement with only a few lines of code.

The contributions of this paper are summarized as follows. First, to facilitate data augmentation for domain generalization, we for the first time introduce a new concept called MixStyle, which builds a connection between visual domains and feature statistics that control image styles. Second, we provide an efficient implementation of MixStyle through simple manipulation of feature statistics. Third, a MixStyle-based semi-supervised learning framework is proposed to extend the applicability of MixStyle to partially labeled datasets. Finally, extensive experiments are conducted to demonstrate the effectiveness and versatility of MixStyle in the problems of domain generalization, semi-supervised domain generalization, and unsupervised domain adaptation, and covering various tasks including object recognition, instance retrieval, and RL. Comprehensive ablation studies are also provided to give an in-depth understanding into the mechanism of MixStyle, as well as to share insights on how to apply MixStyle in practice. All source code of this work has been made publicly available to facilitate future research.\footnote{\url{https://github.com/KaiyangZhou/mixstyle-release}}

An earlier and preliminary study of MixStyle was published in ICLR 2021~\cite{zhou2021mixstyle}. In comparison, this paper introduces substantial new materials: 1) MixStyle can now work with unlabeled data via a non-trivial design that allows the mixing of feature statistics between labeled and pseudo-labeled instances; 2) We demonstrate that MixStyle works exceptionally well with extremely limited labels like 5 labels per category; 3) Extensive experiments covering semi-supervised domain generalization~\cite{zhou2021stylematch} and unsupervised domain adaptation~\cite{ganin2015unsupervised} are conducted.

\section{Related Work}

\minisec{Domain Generalization (DG)}
studies the problem of model generalization to out-of-distribution data using only source data that are gathered from multiple relevant domains. We refer readers to a recent survey paper~\cite{zhou2021domain} for a more comprehensive literature review in this topic. Numerous methods are based on the idea of aligning feature distributions across different source domains. The so-called domain alignment methods often minimize a distance metric that quantifies the source domain discrepancy, such as those based on maximum mean discrepancy (MMD)~\cite{li2018mmdaae}, contrastive losses~\cite{motiian2017unified}, and adversarial learning~\cite{li2018ciddg}. Since source data cover multiple distinct domains, ensemble learning has also been explored where the main idea is to learn domain-specific models, such as domain-specific classifiers~\cite{zhou2020domain,ding2017deep} or batch normalization layers~\cite{seo2020learning,segu2023batch}, and use their ensemble for prediction. Recently, meta-learning has drawn increasing attention from the DG community~\cite{li2018learning,balaji2018metareg,dou2019domain}. The key idea is to construct training episodes that are fed to the model at each forward-backward step. Each episode contains a pseudo-train and a pseudo-test set with non-overlapping domains, both derived from source domains. The model is typically optimized in the pseudo-train set in a way that the performance in the pseudo-test set is improved, which often requires second-order differentiation.

More related to our work are data augmentation methods. CrossGrad~\cite{shankar2018generalizing} perturbs the input to a category classifier with adversarial gradients back-propagated from a domain classifier. DDAIG~\cite{zhou2020deep} learns a perturbation neural network to synthesize images that cannot be recognized by a domain classifier. L2A-OT~\cite{zhou2020learning} learns a neural network to map source data to pseudo-novel domains by maximizing an optimal transport-based distance measure. The proposed MixStyle is related to L2A-OT in its effort to synthesizing novel domains. However, MixStyle differs from L2A-OT in its much simpler formulation that leverages feature statistics, as well as its more efficient implementation that requires only a few lines of code. Essentially, MixStyle can be seen as feature-level augmentation, which is clearly different from image-level augmentation methods like L2A-OT.

\minisec{Generalization in Deep RL}
Reinforcement learning (RL) agents often overfit training environments and as a result perform poorly in unseen environments with different visual patterns or difficulty levels~\cite{zhang2018study}. A natural way to improve generalization, which has been shown effective in several studies~\cite{cobbe2019quantifying,farebrother2018generalization}, is to impose regularization, such as weight decay. However, Igl et al.~\cite{igl2019generalization} suggest that stochastic regularization methods like dropout~\cite{srivastava2014dropout} and batch normalization~\cite{ioffe2015batch}---the latter relies on estimated population statistics---can cause some adverse effects because the training data in RL are essentially model-dependent. They propose selective noise injection (SNI) that combines a stochastic regularization method with its deterministic counterpart. They further integrate SNI with information bottleneck actor critic (IBAC-SNI) to reduce the variance in gradients. Justesen et al.~\cite{justesen2018illuminating} design a curriculum learning scheme where the level of training episodes progresses from easy to difficult during the course of training. Advances in image-to-image translation have also been exploited, e.g., Gamrian and Goldberg~\cite{gamrian2019transfer} use an image translation model to map target data to the source domain where the agent was trained on to solve the domain shift problem. Tobin et al.~\cite{tobin2017domain} introduce domain randomization, which diversifies training data by rendering images with different visual effects via a programmable simulator. Lee et al.~\cite{lee2020network} pre-process input images with a randomly initialized network for data augmentation. A couple of recent studies~\cite{laskin2020reinforcement,kostrikov2021image} have shown that it is useful to combine a diverse set of label-preserving transformations, such as rotation, shifting and Cutout~\cite{devries2017cutout}. MixStyle is related to the idea of domain randomization but does not need specialized simulators to achieve domain augmentation. MixStyle works efficiently at feature level and is orthogonal to most existing methods, such as IBAC-SNI (see Section~\ref{sec:exp;ssec:dg;sssec:rl}), as it can be simply plugged into RL agents' CNN backbone.

\begin{figure*}[t]
    \centering
    \includegraphics[width=\textwidth]{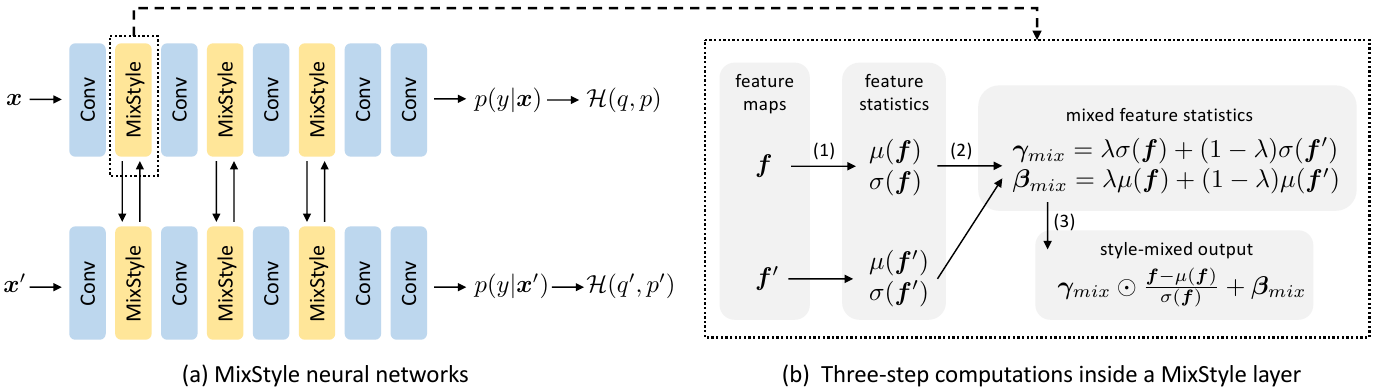}
    \caption{A schematic overview of MixStyle neural networks. The best practice is to insert MixStyle to multiple shallow layers. $\bm{x}$ and $\bm{x}'$ denote two image instances, with $q$ and $q'$ being their ground-truth labels (or one of them being pseudo label if dealing with unlabeled data). $\mathcal{H}(\cdot, \cdot)$ means the cross-entropy loss. $\bm{f}$ and $\bm{f}'$ are feature maps extracted at a certain layer for $\bm{x}$ and $\bm{x}'$, respectively. $\mu(\cdot)$ and $\sigma(\cdot)$ represent functions for computing channel-wise means and standard deviations, respectively. $\lambda$ is an instance-specific, random weight sampled from the beta distribution.}
    \label{fig:method}
\end{figure*}

\minisec{Unsupervised Domain Adaptation (UDA)}
is closely related to DG because both aim to overcome the domain shift problem encountered in a new target domain. The biggest difference lies in whether target data are available for training: the former assumes unlabeled target data are available whereas the latter relies purely on source data. A large body of UDA methods~\cite{ganin2015unsupervised,sun2016deep,tzeng2017adversarial,long2016unsupervised,deng2019cluster,kang2020contrastive,saito2018maximum,liu2020open} fall into the group of domain alignment, which sounds familiar in the DG literature. However, in UDA this idea is backed by a theoretical guarantee~\cite{ben2010theory} that reducing the distribution mismatch between the source and target domain can lead to a lower upper bound on the target error. Methods fallen into this group often seek to minimize an explicit distance function based on, e.g., moments~\cite{sun2016deep} or MMD~\cite{long2016unsupervised}, or learn a domain discriminator in an adversarial manner that equivalently minimizes the Jensen-Shannon divergence~\cite{ganin2015unsupervised,tzeng2017adversarial}. Recent studies~\cite{deng2019cluster,saito2018maximum,kang2020contrastive} have suggested that fine-grained, class-specific alignment outperforms coarse alignment---the latter might result in misalignment in samples of different classes. Besides feature-level alignment, pixel-level alignment has also been extensively researched where the key idea is to transform source images into the target style so a target classifier can be directly trained~\cite{hoffman2018cycada,gong2018dlow}. A related work is DLOW~\cite{gong2018dlow}, which maps source images to intermediate domains as well as the target domain to achieve model adaptation for semantic segmentation. When applied to UDA, our MixStyle mixes source and target feature statistics, which can be seen as generating intermediate domains, like DLOW---but in an implicit yet much more efficient way.

\section{MixStyle Neural Networks}
Generalization to unseen domains using only source data for training is a challenging task. Our motivation to improve out-of-distribution generalization for neural networks is to diversify the source data distributions (domains) as much as possible. This allows the model to discover more generalizable patterns by itself through learning. However, manually collecting data from more sources for data augmentation is both expensive and time-consuming. To achieve data augmentation in a more efficient way, we resort to feature-level augmentation.

Our proposed approach, \emph{MixStyle}, is a plug-and-play CNN module that probabilistically mixes the CNN feature statistics between instances to implicitly synthesize ``new'' styles. This is inspired by recent studies~\cite{ulyanov2016instance,huang2017arbitrary,dumoulin2017learned} showing that CNN feature statistics---specifically the channel-wise means and standard deviations of feature maps---essentially capture the style of an image, which can be viewed as a characterization of visual domains.

A schematic overview of MixStyle neural networks is shown in Fig.~\ref{fig:method}. It is worth noting that our MixStyle module is totally parameter-free and does not need to store any buffer, hence being highly efficient, as well as easy to implement---it only requires a few lines of code using existing open-source deep learning frameworks like PyTorch or TensorFlow. Moreover, there is no need to modify the learning objectives. One can follow the original training paradigm, such as using the cross-entropy loss for image classification tasks or policy gradient methods for reinforcement learning.

Below we introduce in more detail the background on CNN feature statistics, the design of MixStyle module, and finally how MixStyle can be extended to cope with unlabeled data for semi-supervised learning.

\subsection{Background}
\minisec{Instance Normalization}
Most recent style transfer models are based on feedforward neural networks with an encoder-decoder architecture~\cite{huang2017arbitrary,dumoulin2017learned}. Ulyanov et al.~\cite{ulyanov2016instance} identify that replacing batch normalization in feedforward style transfer models with instance normalization can effectively remove instance-specific styles, facilitating image style transfer. The basic idea of instance normalization is to normalize feature maps with means and standard deviations computed across the spatial dimension within each feature channel, followed by a learnable affine transformation. Formally, let $\bm{f} \in \mathbb{R}^{C \times H \times W}$ be feature maps for an image (extracted at a certain CNN layer), with $C$, $H$ and $W$ denoting the dimension of channel, height and width, respectively, instance normalization (IN) is defined as
\begin{equation} \label{eq:instance_norm}
\operatorname{IN}(\bm{f}) = \bm{\gamma} \odot \frac{\bm{f} - \mu(\bm{f})}{\sigma(\bm{f})} + \bm{\beta},
\end{equation}
where $\bm{\gamma}, \bm{\beta} \in \mathbb{R}^C$ denote the affine transformation parameters; $\mu(\bm{f}), \sigma(\bm{f}) \in \mathbb{R}^C$ are the means and standard deviations computed within each channel of $\bm{f}$. Specifically, for $c \in \{1, ..., C\}$,
\begin{equation} \label{eq:instance_mean}
\mu(\bm{f})_c = \frac{1}{HW} \sum_{h=1}^H \sum_{w=1}^W f_{c,h,w},
\end{equation}
and
\begin{equation} \label{eq:instance_std}
\sigma(\bm{f})_c = \sqrt{ \frac{1}{HW} \sum_{h=1}^H \sum_{w=1}^W ( f_{c,h,w} - \mu(\bm{f})_c )^2 }.
\end{equation}

\minisec{Adaptive Instance Normalization}
Since the feature statistics are linked to image style, Huang and Belongie~\cite{huang2017arbitrary} propose adaptive instance normalization (AdaIN) that replaces the feature statistics of a content image with those of a style image to achieve arbitrary image style transfer. Specifically, AdaIN replaces the affine transformation parameters, $\bm{\gamma}$ and $\bm{\beta}$ in Eq.~\eqref{eq:instance_norm}, with the feature statistics of a style image, $\mu(\bm{f}')$ and $\sigma(\bm{f}')$, where $\bm{f}'$ denotes the style image's feature maps,
\begin{equation} \label{eq:adain}
\operatorname{AdaIN}(\bm{f}, \bm{f}') = \sigma(\bm{f}') \odot \frac{\bm{f} - \mu(\bm{f})}{\sigma(\bm{f})} + \mu(\bm{f}').
\end{equation}

\subsection{MixStyle Module}
MixStyle draws inspiration from AdaIN. However, rather than attaching a decoder for image generation, MixStyle is designed for the purpose of regularizing CNN training by perturbing the style information of source domain instances. MixStyle is inserted between layers (blocks) in a CNN architecture, as shown in Fig.~\ref{fig:method}(a).

More specifically, MixStyle mixes the feature statistics of two instances with a random convex weight. The computations inside a MixStyle module can be summarized into three steps (depicted in Fig.~\ref{fig:method}(b)). First, given two sets of feature maps $\bm{f}$ and $\bm{f}'$ for two instances, MixStyle computes their feature statistics, $(\mu(\bm{f}), \sigma(\bm{f}))$ and $(\mu(\bm{f}'), \sigma(\bm{f}'))$. Second, MixStyle generates a mixture of feature statistics,
\begin{align}
\bm{\gamma}_{mix} &= \lambda \sigma(\bm{f}) + (1 - \lambda) \sigma(\bm{f}'), \label{eq:mix_gamma} \\
\bm{\beta}_{mix} &= \lambda \mu(\bm{f}) + (1 - \lambda) \mu(\bm{f}'), \label{eq:mix_beta}
\end{align}
where $\lambda$ is an instance-specific, random weight sampled from the beta distribution, $\lambda \sim Beta(\alpha, \alpha)$ with $\alpha \in (0, \infty)$ being a hyper-parameter. We suggest setting $\alpha = 0.1$ in practice.

Finally, the mixture of feature statistics is applied to the style-normalized $\bm{f}$,
\begin{equation} \label{eq:mixstyle}
\operatorname{MixStyle}(\bm{f}, \bm{f}') = \bm{\gamma}_{mix} \odot \frac{\bm{f} - \mu(\bm{f})}{\sigma(\bm{f})} + \bm{\beta}_{mix}.
\end{equation}

Essentially, AdaIN is a special case of MixStyle when $\lambda = 0$. We use a probability of 0.5 to decide if MixStyle is activated or not in the forward pass. MixStyle is \emph{not} used at test time. Gradients are blocked in the computational graph of $\mu(\cdot)$ and $\sigma(\cdot)$ to prevent the augmentation effect from being erased.

\minisec{Mini-Batch Training}
MixStyle can be easily integrated into mini-batch training. Specifically, $\operatorname{MixStyle}(\bm{f}, \bm{f}')$ in Eq.~\eqref{eq:mixstyle} can be modified to $\operatorname{MixStyle}(\bm{F}, \bm{F}')$, with $\bm{F}, \bm{F}' \in \mathbb{R}^{B \times C \times H \times W}$ being two batches of feature maps ($B$ denotes the batch size). A simple realization is to shuffle the order in the batch dimension of $\bm{F}$ to obtain $\bm{F}'$. When source domain labels are available, one can also force style mixing to occur between images of different domains. This can be implemented by sampling half of the $B$ images from one source domain while the other half from a different source domain, and switching the order of the two halves in $\bm{F}$ to produce $\bm{F}'$. Fig.~\ref{fig:motivation} suggests that sub-domains exist within each domain, so even if two instances of the same domain are mixed, new domains could still be synthesized. We empirically find that the performance of random mixing is comparable to that of cross-domain mixing.

\begin{table*}[t]
\tabstyle{5pt}
\caption{Domain generalization results on PACS and Office-Home. MixStyle outperforms other strong regularization methods, such as DropBlock and CutMix, by a clear margin. MixStyle also achieves comparable performance with the main DG competitor, L2A-OT, which synthesizes novel-domain data in the input space---MixStyle achieves implicit data augmentation in the feature space, which is much more efficient.}
\label{tab:result_dg}
\begin{tabular}{l cccc >{\em}c cccc >{\em}c}
\toprule
\multirow{2}{*}{Model} & \multicolumn{5}{c}{PACS} & \multicolumn{5}{c}{Office-Home} \\
\cmidrule(lr){2-6}\cmidrule(lr){7-11}
& Art painting & Cartoon & Photo & Sketch & Avg & Art & Clipart & Product & Real-world & Avg \\
\midrule
MMD-AAE~\cite{li2018mmdaae} & 75.2 & 72.7 & 96.0 & 64.2 & 77.0 & 56.5 & 47.3 & 72.1 & 74.8 & 62.7 \\
CCSA~\cite{motiian2017unified} & 80.5 & 76.9 & 93.6 & 66.8 & 79.4 & 59.9 & 49.9 & 74.1 & 75.7 & 64.9 \\
JiGen~\cite{cvpr19jigen} & 79.4 & 75.3 & 96.0 & 71.6 & 80.5 & 53.0 & 47.5 & 71.5 & 72.8 & 61.2 \\
CrossGrad~\cite{shankar2018generalizing} & 79.8 & 76.8 & 96.0 & 70.2 & 80.7 & 58.4 & 49.4 & 73.9 & 75.8 & 64.4 \\
Epi-FCR~\cite{li2019episodic} & 82.1 & 77.0 & 93.9 & 73.0 & 81.5 & - & - & - & - & - \\
Metareg~\cite{balaji2018metareg} & 83.7 & 77.2 & 95.5 & 70.3 & 81.7 & - & - & - & - & - \\
L2A-OT~\cite{zhou2020learning} & {83.3} & {78.2} & {96.2} & {73.6} & {82.8} & \textbf{60.6} & 50.1 & \textbf{74.8} & \textbf{77.0} & \textbf{65.6} \\
\midrule
ResNet-18 & 77.0 & 75.9 & 96.0 & 69.2 & 79.5 & 58.9 & 49.4 & 74.3 & 76.2 & 64.7 \\
+ Manifold Mixup~\cite{verma2019manifold} & 75.6 & 70.1 & 93.5 & 65.4 & 76.2 & 56.2 & 46.3 & 73.6 & 75.2 & 62.8 \\
+ Cutout~\cite{devries2017cutout} & 74.9 & 74.9 & 95.9 & 67.7 & 78.3 & 57.8 & 48.1 & 73.9 & 75.8 & 63.9 \\
+ CutMix~\cite{yun2019cutmix} & 74.6 & 71.8 & 95.6 & 65.3 & 76.8 & 57.9 & 48.3 & 74.5 & 75.6 & 64.1 \\
+ Mixup~\cite{zhang2018mixup} & 76.8 & 74.9 & 95.8 & 66.6 & 78.5 & 58.2 & 49.3 & 74.7 & 76.1 & 64.6 \\
+ DropBlock~\cite{ghiasi2018dropblock} & 76.4 & 75.4 & 95.9 & 69.0 & 79.2 & 58.0 & 48.1 & 74.3 & 75.9 & 64.1 \\
+ MixStyle (random) & {82.3} & \textbf{79.0} & \textbf{96.3} & {73.8} & {82.8} & 58.7 & \textbf{53.4} & 74.2 & 75.9 & 65.5 \\
+ MixStyle (xdomain) & \textbf{84.1} & 78.8 & 96.1 & \textbf{75.9} & \textbf{83.7} & 57.3 & 52.9 & 73.5 & 75.3 & 64.8 \\
\bottomrule
\end{tabular}
\end{table*}

\subsection{Extension to Semi-Supervised Learning}
We present a simple MixStyle-based semi-supervised learning framework to deal with unlabeled data. This new framework can be applied to both semi-supervised domain generalization~\cite{zhou2021stylematch}, where the source data are only partially labeled, and unsupervised domain adaptation~\cite{lu2020stochastic,saito2018maximum}, where the goal is to adapt a model from a labeled source dataset to an unlabeled target dataset. Compared to supervised learning, the semi-supervised version mainly differs in that two to-be-mixed instances (or mini-batches) come from the labeled and the unlabeled dataset respectively. More specifically, $\bm{x}$ in Fig.~\ref{fig:method}(a) is a labeled instance while $\bm{x}'$ is now an unlabeled instance. For semi-supervised domain generalization, $\bm{x}'$ is sampled from the unlabeled source dataset, whereas for unsupervised domain adaptation, $\bm{x}'$ comes from the unlabeled target domain.

Since the MixStyle training requires proper supervision for learning meaningful representations, we propose to assign pseudo labels to the unlabeled data using predictions from the model itself. For each unlabeled instance $\bm{x}'$, the model produces a probability distribution over all categories, and the one with the highest confidence estimate is selected as the pseudo label. Following FixMatch~\cite{sohn2020fixmatch}, a state-of-the-art pseudo-labeling method, we use a confidence threshold to filter out low-confidence predictions, and adopt the weak-strong augmentation training to reduce overfitting to noisy pseudo labels---pseudo labels are estimated on weakly augmented images while predictions are made on strongly augmented images.

Formally, given a batch of $B$ instances---half of them are labeled and half of them are unlabeled---we have a supervised loss $\ell_s$, which is computed on the labeled instances,
\begin{equation} \label{eq:loss_s}
\ell_s = \sum_{i=1}^{\frac{B}{2}} \mathcal{H}(q_i, p(y | a(\bm{x}_i))),
\end{equation}
where $\mathcal{H}(\cdot, \cdot)$ is the cross-entropy loss, $q_i$ is the ground-truth label of $\bm{x}_i$, and $a(\cdot)$ denotes a weak augmentation function (e.g., random flip and crop).

We also have an unlabeled loss $\ell_u$, which is computed on the unlabeled instances,
\begin{equation} \label{eq:loss_u}
\ell_u = \sum_{i=1}^{\frac{B}{2}} \mathbbm{1}(\max(p'_i) \geq \tau) \mathcal{H}(q'_i, p(y | A(\bm{x}'_i))),
\end{equation}
where $\max(p'_i)$ is the most confident estimate, $\tau$ is the confidence threshold (fixed to 0.95), $q'_i$ is the pseudo label obtained on the weakly augmented image $a(\bm{x}'_i)$, and $A(\cdot)$ denotes a strong augmentation function (e.g., RandAugment~\cite{cubuk2019randaugment}).

\begin{table}[t]
    \tabstyle{6pt}
    \caption{Results on DomainBed. (Model selection: training-domain validation set.)}
    \label{tab:result_domainbed}
    \begin{tabular}{l ccc}
    \toprule
    & PACS & VLCS & Office-Home \\
    \midrule
    ERM & 84.2 & 77.3 & 67.6 \\
    MixStyle & 85.2 & 77.9 & 60.4 \\
    \bottomrule
    \end{tabular}
\end{table}

\begin{figure}[t]
\centering
\includegraphics[width=\columnwidth]{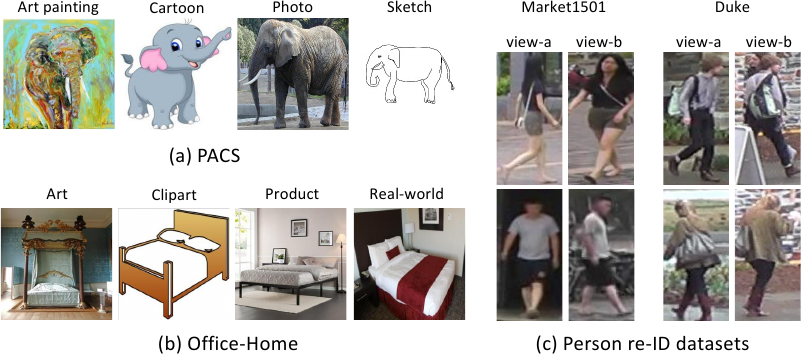}
\caption{Example images from PACS, Office-Home and the person re-ID benchmark.}
\label{fig:im_pacs_oh_reid}
\end{figure}

\begin{table*}[t]
\tabstyle{8pt}
\caption{Results on the cross-dataset person re-identification (re-ID) benchmark. MixStyle demonstrates clear advantages over the baseline methods in all settings. It is worth noting that the domain shift problem here is much more challenging than that in the object recognition benchmarks since the person images are captured by real-world cameras deployed in complex scenes.}
\label{tab:result_reid}
\begin{tabular}{l cccc cccc}
\toprule
\multirow{2}{*}{Model} & \multicolumn{4}{c}{Market1501$\to$Duke} & \multicolumn{4}{c}{Duke$\to$Market1501} \\
\cmidrule(lr){2-5}\cmidrule(lr){6-9}
& mAP & R1 & R5 & R10 & mAP & R1 & R5 & R10 \\
\midrule
ResNet-50 & 19.3 & 35.4	& 50.3 & 56.4 & 20.4 & 45.2 & 63.6 & 70.9 \\
+ RandomErase~\cite{zhong2020random} & 14.3 & 27.8 & 42.6 & 49.1 & 16.1 & 38.5 & 56.8 & 64.5 \\
+ DropBlock~\cite{ghiasi2018dropblock} & 18.2 & 33.2 & 49.1 & 56.3 & 19.7 & 45.3 & 62.1 & 69.1 \\
+ MixStyle (random) & \textbf{23.8} & {42.2} & {58.8} & \textbf{64.8} & {24.1} & {51.5} & {69.4} & {76.2} \\
+ MixStyle (xdomain) & 23.4 & \textbf{43.3} & \textbf{58.9} & 64.7 & \textbf{24.7} & \textbf{53.0} & \textbf{70.9} & \textbf{77.8} \\
\midrule
OSNet & 25.9 & 44.7 & 59.6 & 65.4 & 24.0 & 52.2 & 67.5 & 74.7 \\
+ RandomErase~\cite{zhong2020random} & 20.5 & 36.2 & 52.3 & 59.3 & 22.4 & 49.1 & 66.1 & 73.0 \\
+ DropBlock~\cite{ghiasi2018dropblock} & 23.1 & 41.5 & 56.5 & 62.5 & 21.7 & 48.2 & 65.4 & 71.3 \\
+ MixStyle (random) & {27.2} & \textbf{48.2} & \textbf{62.7} & \textbf{68.4} & {27.8} & {58.1} & {74.0} & \textbf{81.0} \\
+ MixStyle (xdomain) & \textbf{27.3} & 47.5 & 62.0 & 67.1 & \textbf{29.0} & \textbf{58.2} & \textbf{74.9} & 80.9 \\
\bottomrule
\end{tabular}
\end{table*}

\begin{figure*}[t]
    \centering
    \includegraphics[width=.95\textwidth]{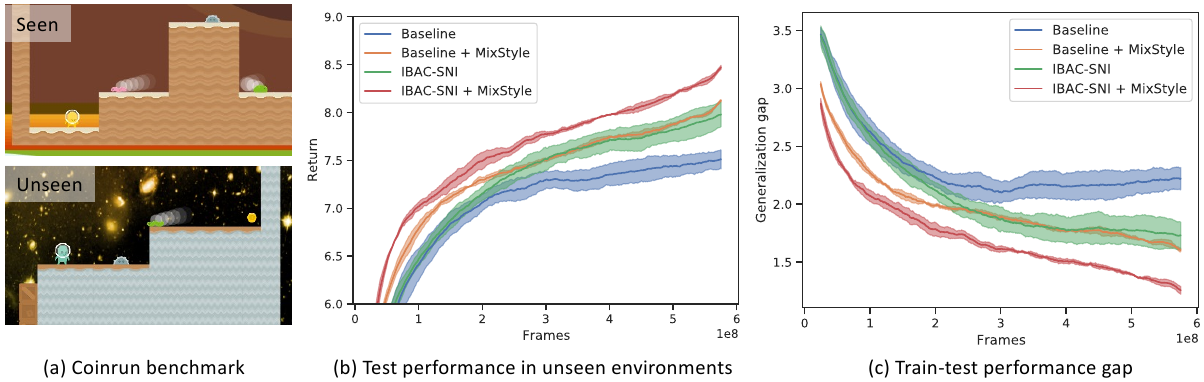}
    \caption{Generalization results on the Coinrun benchmark. MixStyle significantly boosts the generalization performance of both the baseline model and the state-of-the-art method, IBAC-SNI.}
    \label{fig:coinrun_main}
\end{figure*}

\section{Experiments} \label{sec:exp}

\subsection{Domain Generalization} \label{sec:exp;ssec:dg}
To demonstrate the versatility of MixStyle, our experiments cover a wide range of generalization tasks: object recognition (Section~\ref{sec:exp;ssec:dg;sssec:object}), instance retrieval (Section~\ref{sec:exp;ssec:dg;sssec:reid}), and reinforcement learning (Section~\ref{sec:exp;ssec:dg;sssec:rl}).

\subsubsection{Object Recognition} \label{sec:exp;ssec:dg;sssec:object}
\minisec{Experimental Setup}
We conduct experiments on two commonly used domain generalization (DG) datasets, namely PACS~\cite{li2017deeper} and Office-Home~\cite{office_home}, which contain different types of domain shifts. PACS consists of four domains---art painting, cartoon, photo and sketch---with a total of 9,991 images and seven classes. The domain shifts are mainly caused by image style changes (Fig.~\ref{fig:im_pacs_oh_reid}(a)). Office-Home also comprises four domains, which are art, clipart, product and real-world, but contains more images than PACS: around 15,500 images of 65 classes related to objects in office and home environments (Fig.~\ref{fig:im_pacs_oh_reid}(b)). Following previous work~\cite{li2019episodic,zhou2020learning}, we use the leave-one-domain-out protocol for evaluation. Specifically, we pick three domains as the source to train a model, which is then deployed in the remaining domain seen as the target. All experiments are repeated five times with different random seeds and the average results are reported.

\minisec{Implementation Details}
We adopt the commonly used ImageNet-pretrained ResNet-18~\cite{he2016deep} as the CNN backbone, same as existing methods~\cite{cvpr19jigen,li2019episodic,zhou2020learning}. We set the batch size to 64 and train the model using the SGD optimizer for 50 epochs. The initial learning rate is set to 0.001 and decayed by the cosine annealing rule~\cite{loshchilov2016sgdr}. For PACS, MixStyle is inserted after the 1st, 2nd and 3rd residual blocks. For Office-Home, MixStyle is inserted after the 1st and 2nd residual blocks. The study of where to insert MixStyle in a CNN will later be presented in Section~\ref{sec:exp;ssec:ablation}. Our code is built using the public \texttt{Dassl.pytorch} toolbox~\cite{zhou2020domain}.\footnote{\url{https://github.com/KaiyangZhou/Dassl.pytorch}}

\minisec{Baselines}
Our main competitors are general-purpose regularization methods including Mixup~\cite{zhang2018mixup}, Manifold Mixup~\cite{verma2019manifold}, DropBlock~\cite{ghiasi2018dropblock}, CutMix~\cite{yun2019cutmix}, and Cutout~\cite{devries2017improved}. They are trained using the same optimization parameters as MixStyle, along with the optimal hyper-parameters setup as suggested in their papers. We also compare with the existing DG methods, which reported state-of-the-art (SOTA) results on PACS and Office-Home. These methods include the domain alignment-based CCSA~\cite{motiian2017unified} and MMD-AAE~\cite{li2018mmdaae}, the Jigsaw puzzle-based JiGen~\cite{cvpr19jigen}, the adversarial gradient-based CrossGrad~\cite{shankar2018generalizing}, the meta-learning-based Metareg~\cite{balaji2018metareg} and Epi-FCR~\cite{li2019episodic}, and the data augmentation-based L2A-OT~\cite{zhou2020learning}.

\minisec{Results}
The comprehensive results are presented in Table~\ref{tab:result_dg}. We first discuss the comparisons with the general-purpose regularization methods. Overall, we observe that the general-purpose regularization methods do not offer any clear advantage over the vanilla ResNet-18 in these DG tasks. In contrast, MixStyle (both the random and cross-domain mixing versions) consistently improves upon the vanilla model on both datasets. Compared with Mixup, which mixes images in the pixel space, our feature statistics-based MixStyle is around 4\% better on PACS and 1\% better on Office-Home. MixStyle also beats Manifold Mixup, which further justifies the advantage of mixing feature statistics for DG. MixStyle and DropBlock share some commonalities in that both are applied to feature maps at multiple layers, but MixStyle shows clear advantages over DropBlock in all domains. The reason why DropBlock is ineffective here is because dropping out feature activations mainly encourages the network to mine discriminative patterns but does not reinforce the ability to cope with unseen domains (styles), which is however what MixStyle aims to achieve through synthesizing ``new'' styles for data augmentation. In addition, it is interesting to see that the random mixing version of MixStyle is highly comparable to the cross-domain mixing version---the former achieves slightly better performance on two domains of PACS (cartoon and photo) and all domains of Office-Home. These results suggest that there indeed exist sub-domains within a single source domain, which allow random mixing to produce more diverse domains, and hence a more domain-generalizable model.

When it comes to SOTA DG methods, MixStyle outperforms most of them by a clear margin, despite having a much simpler design that adds negligible overhead to a supervised classifier. On PACS, the cross-domain mixing version of MixStyle even surpasses the recently introduced L2A-OT, with nearly 1\% improvement on average. From a data augmentation perspective, MixStyle and L2A-OT share a similar goal: to synthesize data from pseudo-novel domains. However, the ways they achieve the goal are distinguishable: MixStyle mixes feature statistics in each forward pass, which is highly efficient; whereas L2A-OT trains a deep image generation network by iteratively maximizing the domain difference measured by optimal transport between the original and generated images. The latter introduces much heavier computational cost than MixStyle in both GPU memories and training time, and also requires more engineering effort.

\minisec{DomainBed}
We also conduct experiments on the DomainBed benchmark~\cite{gulrajani2020search} where a larger model is used, i.e., ResNet50, and each method undergoes an extensive parameter tuning process. We compare MixStyle with the ERM model and show the results on Table~\ref{tab:result_domainbed}. Overall, we observe that with the model becoming larger, MixStyle's advantage over ERM shrinks, which is similar to other DG methods~\cite{gulrajani2020search}. On PACS and VLCS, MixStyle beats ERM. But on OfficeHome, MixStyle underperforms ERM by a large margin.

\subsubsection{Instance Retrieval} \label{sec:exp;ssec:dg;sssec:reid}

\minisec{Experimental Setup}
We evaluate MixStyle on the person re-identification (re-ID) task, which aims to match people across disjoint camera views. As each camera view is itself a distinct domain, person re-ID is essentially a cross-domain image (instance) matching problem. We select two datasets for evaluation, namely Market1501~\cite{zheng2015scalable} and Duke~\cite{ristani2016performance,zheng2017unlabeled}. Market1501 contains 32,668 images of 1,501 identities captured by 6 cameras, while Duke contains 36,411 images of 1,812 identities captured by 8 cameras. See Fig.~\ref{fig:im_pacs_oh_reid}(c) for image examples. A model is trained using one dataset and tested on the other.\footnote{We follow the original train/test split in each dataset for evaluation.} Images from these two datasets manifest substantial differences in terms of backgrounds, illuminations, image resolutions, viewpoints, and so on, which greatly challenge cross-domain recognition. Ranking accuracy and mean average precision (mAP)~\cite{zheng2015scalable} are used as the performance measures.

\minisec{Implementation Details}
We evaluate MixStyle using two different CNN architectures: ResNet-50~\cite{he2016deep} and OSNet~\cite{zhou2021osnet}. The latter was specifically designed for re-ID. In both architectures, MixStyle is inserted after the 1st and 2nd residual blocks. Following Zhou et al.~\cite{zhou2021osnet}, we train the model to perform classification on training identities and use features extracted from the penultimate layer (before the linear classifier) for image matching. SGD is used as the optimizer. The batch size is set to 32 and the total number of epochs is 60. The global learning rate is set to 0.05 and decayed by 0.1 every 20 epochs. A smaller learning rate ($0.2\times$ the global learning rate) is applied to the pretrained layers. Our code is built using the \texttt{Torchreid} library~\cite{torchreid}.\footnote{\url{https://github.com/KaiyangZhou/deep-person-reid}}

\minisec{Baselines}
We compare with three baseline methods: 1) The vanilla model, which serves as a strong baseline; 2) DropBlock, which was the top-performing competitor in the object recognition experiment; 3) RandomErase~\cite{zhong2020random}, a widely used regularization method in the re-ID literature (similar to Cutout).

\minisec{Results}
The results are reported in Table~\ref{tab:result_reid}. It is encouraging that MixStyle consistently outperforms the strong vanilla model under both settings and with different architectures. This strongly demonstrates the generality of MixStyle. Again, the results of the random and cross-domain mixing are generally comparable to each other, suggesting that domain labels are indeed not a must. In contrast, DropBlock and RandomErase are unable to show any benefit. Notably, RandomErase, which simulates occlusion by erasing pixels in random rectangular regions with random values, has been used as a default trick when training re-ID models. However, RandomErase shows a detrimental effect in the cross-dataset setting, suggesting that a further investigation into the use of data augmentation methods for practical re-ID is required. Similar to DropBlock, randomly erasing pixels offers no guarantee to improve the robustness when it comes to domain shifts.

\subsubsection{Reinforcement Learning} \label{sec:exp;ssec:dg;sssec:rl}

\minisec{Experimental Setup}
We conduct experiments on the Coinrun benchmark\cite{cobbe2019quantifying}, which is a recently introduced RL dataset particularly for evaluating the generalization performance of RL algorithms. As shown in Fig.~\ref{fig:coinrun_main}(a), the goal is to control the character to collect golden coins while avoiding both stationary and dynamic obstacles. Training data are sampled from 500 levels whereas test data are sampled from new levels of only the highest difficulty.

\minisec{Implementation Details}
We follow Igl et al.~\cite{igl2019generalization} to construct and train the RL agent. The CNN architecture used by IMPALA~\cite{impala2018} is adopted as the policy network and trained by the proximal policy optimization algorithm~\cite{schulman2017proximal}. We refer readers to Igl et al.~\cite{igl2019generalization} for more implementation details. MixStyle is inserted after the 1st and 2nd convolutional sequences. As domain labels are difficult to define, we only evaluate the random mixing version of MixStyle. Our code is built on top of Igl et al.~\cite{igl2019generalization}.\footnote{\url{https://github.com/microsoft/IBAC-SNI}}

\minisec{Baselines}
Following Igl et al.~\cite{igl2019generalization}, we train strong baseline models and add MixStyle on top of them to see whether MixStyle can bring further improvements. To this end, we train two baseline models: 1) Baseline, which combines weight decay and data augmentation;\footnote{We do not use batch normalization or dropout because they are detrimental to the performance, as suggested by Igl et al.~\cite{igl2019generalization}.} and 2) IBAC-SNI (the $\lambda = 0.5$ version)~\cite{igl2019generalization}, which is a SOTA method based on selective noise injection.

\minisec{Results}
The test performance in unseen environments is summarized in Fig.~\ref{fig:coinrun_main}(b). Comparing \blue{Baseline} with \orange{Baseline+MixStyle}, we can see that MixStyle brings a remarkable improvement. Interestingly, MixStyle also significantly reduces the variance, as indicated by the smaller shaded area (for both \orange{orange} and \red{red} lines). These results strongly demonstrate the effectiveness of MixStyle in enhancing generalization for RL agents. When it comes to the stronger baseline \green{IBAC-SNI}, \red{IBAC-SNI+MixStyle} is able to further boost the performance, suggesting that MixStyle is complementary to IBAC-SNI. This result also shows the potential of MixStyle as a plug-and-play component to be combined with other advanced RL methods. It is worth noting that \orange{Baseline+MixStyle} itself is already highly competitive with \green{IBAC-SNI}. Fig.~\ref{fig:coinrun_main}(c) plots the generalization gap, from which we observe that the models trained with MixStyle (\orange{orange} \& \red{red} lines) clearly generalize faster and better than those without using MixStyle (\blue{blue} \& \green{green} lines).

\begin{table*}[t]
\tabstyle{8pt}
\caption{Domain generalization results in the low-data regime on PACS, averaged over 5 random splits. A: Art painting. C: Cartoon. P: Photo. S: Sketch. $*$ means the method uses unlabeled source data for training.}
\label{tab:result_ssdg_pacs}
\begin{tabular}{l cccc >{\em}c cccc >{\em}c}
\toprule
\multirow{2}{*}{Model} & \multicolumn{5}{c}{\# labels: 210 (\emph{10 per class})} & \multicolumn{5}{c}{\# labels: 105 (\emph{5 per class})} \\
\cmidrule(lr){2-6}\cmidrule(lr){7-11}
& A & C & P & S & {Avg} & A & C & P & S & {Avg} \\
\midrule
Vanilla & 63.09 & 58.49 & 86.56 & 45.56 & {63.42} & 56.71 & 53.87 & 71.87 & 36.92 & {54.84} \\
Manifold Mixup~\cite{verma2019manifold} & 60.29 & 54.21 & 81.60 & 39.88 & {59.00} & 53.72 & 49.90 & 69.98 & 34.48 & {52.02} \\
Cutout~\cite{devries2017cutout} & 62.15 & 60.09 & 86.97 & 44.65 & {63.46} & 55.48 & 55.97 & 73.04 & 36.16 & {55.16} \\
CutMix~\cite{yun2019cutmix} & 59.32 & 57.28 & 83.27 & 44.07 & {60.98} & 55.38 & 53.49 & 70.83 & 38.42 & {54.53} \\
Mixup~\cite{zhang2018mixup} & 64.28 & 57.92 & 85.79 & 42.69 & {62.67} & 57.71 & 54.33 & 73.01 & 37.16 & {55.56} \\
DropBlock~\cite{ghiasi2018dropblock} & 61.82 & 59.01 & 86.91 & 48.57 & {64.08} & 55.79 & 53.24 & 73.59 & 37.96 & {55.15} \\
CrossGrad~\cite{shankar2018generalizing} & 62.56 & 58.92 & 85.81 & 44.11 & {62.85} & 56.39 & 55.11 & 72.61 & 38.08 & {55.55} \\
DDAIG~\cite{zhou2020deep} & 61.95 & 58.74 & 84.44 & 47.48 & {63.15} & 55.09 & 52.31 & 70.53 & 38.89 & {54.20} \\
MixStyle & \textbf{71.11} & \textbf{64.04} & \textbf{88.99} & \textbf{54.62} & \textbf{69.69} & \textbf{62.00} & \textbf{58.40} & \textbf{80.43} & \textbf{43.58} & \textbf{61.10} \\
\midrule
$^*$EISNet~\cite{wang2020learning} & 66.84 & 61.33 & 89.16 & 51.38 & {67.18} & 62.08 & 54.75 & 80.66 & 42.68 & {60.04} \\
$^*$FixMatch~\cite{sohn2020fixmatch} & 78.01 & 68.93 & 87.79 & 73.75 & {77.12} & 77.30 & 68.67 & 80.49 & 73.32 & {74.94} \\
$^*$StyleMatch~\cite{zhou2021stylematch} & 79.43 & 73.75 & 90.04 & \textbf{78.40} & {80.41} & 78.54 & \textbf{74.44} & 89.25 & \textbf{79.06} & \textbf{80.32} \\
$^*$MixStyle & \textbf{83.89} & \textbf{75.36} & \textbf{92.48} & 73.05 & \textbf{81.19} & \textbf{80.93} & 73.18 & \textbf{90.48} & 70.71 & 78.82 \\
\bottomrule
\end{tabular}
\end{table*}

\begin{table*}[t]
\tabstyle{8pt}
\caption{Domain generalization results in the low-data regime on Office-Home, averaged over 5 random splits. A: Art. C: Clipart. P: Product. R: Real-world. $*$ means the method uses unlabeled source data for training.}
\label{tab:result_ssdg_oh}
\begin{tabular}{l cccc >{\em}c cccc >{\em}c}
\toprule
\multirow{2}{*}{Model} & \multicolumn{5}{c}{\# labels: 1950 (\emph{10 per class})} & \multicolumn{5}{c}{\# labels: 975 (\emph{5 per class})} \\
\cmidrule(lr){2-6}\cmidrule(lr){7-11}
& A & C & P & R & {Avg} & A & C & P & R & {Avg} \\
\midrule
Vanilla & 50.11 & 43.50 & 65.11 & 69.65 & {57.09} & 45.76 & 39.97 & \textbf{60.04} & 63.77 & {52.38} \\
Manifold Mixup~\cite{verma2019manifold} & 47.15 & 39.78 & 63.75 & 67.75 & {54.61} & 44.40 & 37.29 & 58.78 & 62.31 & {50.70} \\
Cutout~\cite{devries2017cutout} & \textbf{50.36} & 43.38 & 65.01 & \textbf{69.86} & {57.15} & 45.70 & 39.63 & 59.67 & 63.70 & {52.18} \\
CutMix~\cite{yun2019cutmix} & 48.95 & 41.48 & 64.05 & 68.31 & {55.70} & 45.32 & 38.40 & 59.00 & 62.49 & {51.30} \\
Mixup~\cite{zhang2018mixup} & 49.67 & 43.87 & 64.83 & 69.12 & {56.87} & 46.29 & 40.30 & 59.02 & 63.87 & {52.37} \\
DropBlock~\cite{ghiasi2018dropblock} & 50.03 & 42.72 & 64.89 & 69.45 & {56.77} & 45.53 & 39.56 & 59.72 & 63.73 & {52.13} \\
CrossGrad~\cite{shankar2018generalizing} & 50.32 & 43.27 & \textbf{65.16} & 69.49 & {57.06} & 45.68 & 40.04 & 59.95 & \textbf{64.09} & {52.44} \\
DDAIG~\cite{zhou2020deep} & 49.60 & 42.52 & 63.54 & 67.89 & {55.89} & 45.73 & 38.82 & 59.52 & 63.37 & {51.86} \\
MixStyle & 49.79 & \textbf{47.12} & 64.18 & 68.42 & \textbf{57.38} & \textbf{46.51} & \textbf{43.59} & 59.66 & 63.30 & \textbf{53.26}\\
\midrule
$^*$EISNet~\cite{wang2020learning} & 51.16 & 43.33 & 64.72 & 68.36 & {56.89} & 47.32 & 40.07 & 59.33 & 62.59 & {52.33} \\
$^*$FixMatch~\cite{sohn2020fixmatch} & 50.36 & 49.70 & 63.93 & 67.56 & {57.89} & 48.98 & 47.46 & 60.70 & 64.36 & {55.38} \\
$^*$StyleMatch~\cite{zhou2021stylematch} & \textbf{52.82} & \textbf{51.60} & \textbf{65.31} & 68.61 & \textbf{59.59} & \textbf{51.53} & \textbf{50.00} & \textbf{60.88} & 64.47 & \textbf{56.72} \\
$^*$MixStyle & 52.44 & 49.61 & 65.01 & \textbf{69.84} & {59.22} & 49.25 & 48.04 & 60.76 & \textbf{64.71} & {55.69} \\
\bottomrule
\end{tabular}
\end{table*}

\subsection{Semi-Supervised Domain Generalization} \label{sec:exp;ssec:ssdg}
Semi-supervised domain generalization (SSDG)~\cite{zhou2021stylematch} considers a more practical scenario where the multi-domain source data are partially labeled. We demonstrate that MixStyle on its own can significantly boost the out-of-distribution generalization performance in the low-data regime with only a few labeled instances per category; and its simple extension that mixes feature statistics between labeled and pseudo-labeled instances shows promising results that rival the current SOTA SSDG methods.

\minisec{Experimental Setup}
We follow the benchmarks designed by Zhou et al.~\cite{zhou2021stylematch}. Specifically, we use PACS and Office-Home (see Section~\ref{sec:exp;ssec:dg;sssec:object} for their statistics). Evaluation is performed on both the 10- and 5-labels-per-class settings (the rest are seen as unlabeled data) and over five random splits.

\minisec{Implementation Details}
The ImageNet-pretrained ResNet-18 is used as the CNN backbone. The configurations for MixStyle are the same as in Section~\ref{sec:exp;ssec:dg;sssec:object}. When training with the labeled data only, we randomly sample a mini-batch of 32 images from a mixture of source domains. Whereas for the semi-supervised extension, we sample 16 images from each source domain to construct a mini-batch (for labeled and unlabeled data respectively). Other training details are kept the same as Zhou et al.~\cite{zhou2021stylematch} for fair comparison.\footnote{\url{https://github.com/KaiyangZhou/ssdg-benchmark}}

\minisec{Baselines}
Similar to the DG experiment on object recognition (Section~\ref{sec:exp;ssec:dg;sssec:object}), we compare with the selected general-purpose regularizers. We also copy the results of other DG and SSDG methods from Zhou et al.~\cite{zhou2021stylematch} for comparison. These include 1) the data augmentation-based CrossGrad~\cite{shankar2018generalizing} and DDAIG~\cite{zhou2020deep}, which can only be trained with fully labeled data. 2) the Jigsaw puzzle-based EISNet~\cite{wang2020learning}, and 3) the latest SSDG method StyleMatch~\cite{zhou2021stylematch}, which is based on a stochastic classifier and image-level style transfer for data augmentation. FixMatch~\cite{sohn2020fixmatch} is also compared here as an ablation study for our semi-supervised MixStyle.

\minisec{Results}
The results on PACS and Office-Home are shown in Table~\ref{tab:result_ssdg_pacs} and~\ref{tab:result_ssdg_oh} respectively. We first discuss the comparisons of methods that only use the labeled source data for training. Compared with the vanilla training, MixStyle brings remarkable improvements on PACS, with over 6\% increase given only 10 labels per class and nearly 7\% increase in the much more challenging 5-labels-per-class setting, both measured by the average accuracy. On Office-Home, the gains are not as large as those on PACS but are still notable given that this dataset poses more challenges due to more classes and cluttered images~\cite{zhou2021stylematch} and that most other methods struggle to beat the vanilla training. We also observe that the general-purpose regularizers and the two data augmentation-based competitors are hugely challenged by the limited labels and are outperformed by MixStyle in most scenarios. When it comes to the comparisons of methods able to utilize the unlabeled source data, MixStyle shows encouraging results that can even rival the latest StyleMatch, despite being much more efficient thanks to the use of feature statistics.


\begin{table}[t]
\tabstyle{10pt}
\caption{Unsupervised domain adaptation results on VisDA-17 using ResNet-101. The purpose here is not to achieve state-of-the-art performance but to show that MixStyle can be integrated into strong baseline methods like FixMatch.}
\label{tab:result_uda_visda}
\begin{tabular}{l c}
\toprule
Model & Accuracy \\
\midrule
Source-only~\cite{saito2018maximum} & 52.4 \\
MMD~\cite{long2015learning} & 61.1 \\
DANN~\cite{ganin2015unsupervised} & 57.4 \\
MCD~\cite{saito2018maximum} & 71.9 \\
SWD~\cite{lee2019sliced} & 76.4 \\
STAR~\cite{lu2020stochastic} & \textbf{82.7} \\
\midrule
FixMatch~\cite{sohn2020fixmatch} & 77.1 \\
MixStyle & 80.0 \\
\bottomrule
\end{tabular}
\end{table}

\begin{table}[t]
\tabstyle{5pt}
\caption{Multi-source unsupervised domain adaptation results on PACS. MixStyle improves upon the strong baseline model, FixMatch, by a clear margin.}
\label{tab:result_msuda_pacs}
\begin{tabular}{l cccc >{\em}c}
\toprule
Model & A & C & P & S & Avg \\
\midrule
Source-only~\cite{wang2020msda} & 75.97 & 73.34 & 91.65 & 64.23 & 76.30 \\
MDAN~\cite{zhao2018adversarial} & 83.54 & 82.34 & 92.91 & 72.42 & 82.80 \\
DCTN~\cite{xu2018deep} & 84.67 & 86.72 & 95.60 & 71.84 & 84.71 \\
M$^3$SDA~\cite{peng2019domainnet} & 84.20 & 85.68 & 94.47 & 74.62 & 84.74 \\
MDDA~\cite{zhao2020multi} & 86.73 & 86.24 & 93.89 & 77.56 & 86.11 \\
LtC-MSDA~\cite{wang2020msda} & 90.19 & 90.47 & 97.23 & 81.53 & 89.85  \\
\midrule
FixMatch~\cite{sohn2020fixmatch} & 90.67 & \textbf{91.56} & 98.83 & 80.15 & 90.30 \\
MixStyle & \textbf{90.77} & 91.05 & \textbf{99.42} & \textbf{88.94} & \textbf{92.55} \\
\bottomrule
\end{tabular}
\end{table}

\begin{figure*}[t]
    \centering
    \includegraphics[width=.95\textwidth]{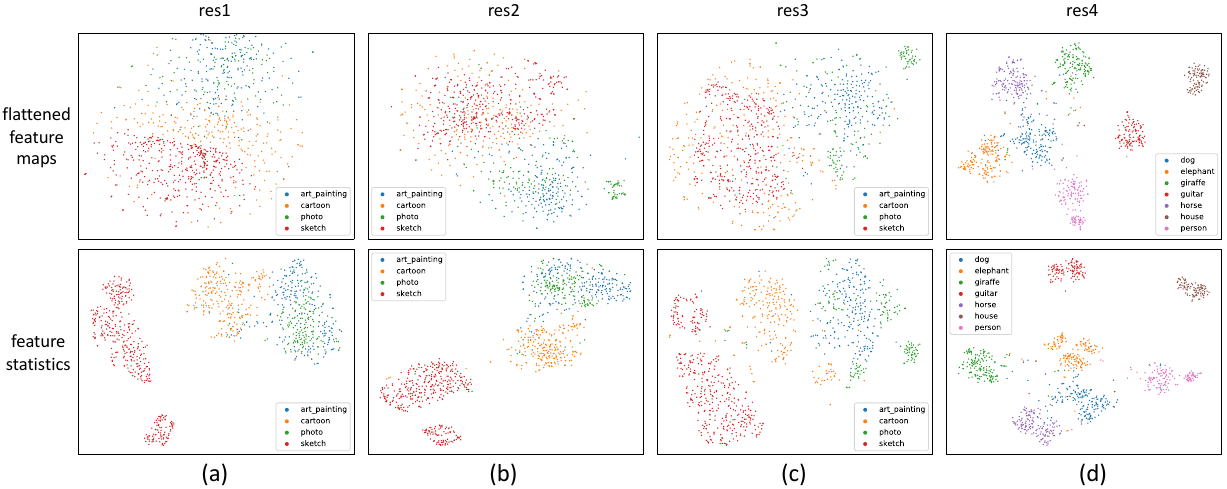}
    \caption{2D visualization of flattened feature maps (top) and the corresponding style statistics (bottom). \texttt{res1-4} denote the four residual blocks in order in a ResNet architecture. We observe that \texttt{res1} to \texttt{res3} contain domain-related information while \texttt{res4} encodes label-related information.}
    \label{fig:tsne_feat_sty}
\end{figure*}

\subsection{Unsupervised Domain Adaptation} \label{sec:exp;ssec:uda}
We further evaluate MixStyle on unsupervised domain adaptation (UDA) where style mixing occurs between the labeled source and pseudo-labeled target images. The experiments cover both single- and multi-source settings. Our goal is not to beat SOTA UDA methods, but to demonstrate that such a simple design based on MixStyle can bring non-trivial improvements.

\minisec{Experimental Setup}
We choose a challenging UDA dataset, VisDA-17~\cite{peng2017visda}, which focuses on knowledge transfer from synthetic to real images. VisDA-17 consists of 152,397 labeled images containing synthetic 3D models and 55,388 unlabeled images derived from MS COCO~\cite{lin2014coco}. There are 12 classes in total. We follow previous work~\cite{saito2018maximum,lu2020stochastic} to report the mean accuracy over 12 classes. For the multi-source setting, we follow Wang et al.~\cite{wang2020msda} to conduct experiments on PACS~\cite{li2017deeper}.

\minisec{Implementation Details}
On VisDA-17, we use the ImageNet-pretrained ResNet-101~\cite{he2016deep} as the CNN backbone and replace the original 1000-way linear classifier with three fully connected layers, following prior work~\cite{saito2018maximum,lu2020stochastic}. MixStyle is inserted after the 1st and 2nd residual blocks. The model is trained with SGD and a constant learning rate of 0.001 for 10 epochs~\cite{saito2018maximum}. The implementation on PACS mostly follows Section~\ref{sec:exp;ssec:ssdg} except that each source mini-batch contains images randomly sampled from the mixture of source domains.

\minisec{Results}
Table~\ref{tab:result_uda_visda} presents the results on VisDA-17.\footnote{Note that for fair comparison we only select the baselines that share a similar implementation including the model architecture.} We observe that the pseudo-labeling-based FixMatch is already a strong UDA model, which easily surpasses most previous specifically designed methods, such as MCD and DANN. Compared with FixMatch, MixStyle gains nearly 3\%, which justifies the importance of the role of cross-domain style mixing. Table~\ref{tab:result_msuda_pacs} shows the results on the multi-source setting where we have similar observations.

\subsection{Ablation Study and Analysis} \label{sec:exp;ssec:ablation}
In this section, we examine the design choices in MixStyle and share insights on how to apply MixStyle in practice.

\begin{table*}[t]
    \tabstyle{5pt}
    \caption{Ablation study on where to apply MixStyle using the ResNet architecture as an example.}
    \label{tab:where_mixstyle}
    \subfloat[Object recognition.]{
        \begin{tabular}{lc}
        \toprule
        Model & Accuracy \\
        \midrule
        ResNet-18 & 79.5 \\
        + MixStyle (\texttt{res1}) & 80.1 \\
        + MixStyle (\texttt{res12}) & 81.6 \\
        + MixStyle (\texttt{res123}) & \textbf{82.8} \\
        + MixStyle (\texttt{res1234}) & 75.6 \\
        + MixStyle (\texttt{res14}) & 76.3 \\
        + MixStyle (\texttt{res23}) & 81.7 \\
        \bottomrule
        \end{tabular}
    }
    ~
    \subfloat[Cross-dataset person re-ID.]{
        \begin{tabular}{lc}
        \toprule
        Model & mAP \\
        \midrule
        ResNet-50 & 19.3 \\
        + MixStyle (\texttt{res1}) & 22.6 \\
        + MixStyle (\texttt{res12}) & \textbf{23.8} \\
        + MixStyle (\texttt{res123}) & 22.0 \\
        + MixStyle (\texttt{res1234}) & 10.2 \\
        + MixStyle (\texttt{res14}) & 11.1 \\
        + MixStyle (\texttt{res23}) & 20.6 \\
        \bottomrule
        \end{tabular}
    }
\end{table*}

\minisec{Where To Apply MixStyle?}
We repeat the DG experiments for object recognition and instance retrieval using the ResNet architecture. Given that a standard ResNet model has four residual blocks denoted by \texttt{res1-4}, we train different models with MixStyle applied to different layers. For notation, \texttt{res1} means MixStyle is applied after the 1st residual block; \texttt{res12} means MixStyle is applied after both the 1st and 2nd residual blocks; and so forth. The results are shown in Table~\ref{tab:where_mixstyle}. We have the following observations. 1) \emph{Applying MixStyle to multiple shallow layers generally achieves a better performance}---for instance, \texttt{res12} is better than \texttt{res1} on both tasks. 2) \emph{Different tasks favor different combinations}---\texttt{res123} achieves the best performance on PACS, while on the re-ID datasets \texttt{res12} is the best. 3) On both tasks, \emph{the performance plunges when applying MixStyle to the last residual block}. This makes sense because \texttt{res4} is the closest to the prediction layer and tends to capture semantic content (i.e.~label-sensitive) information rather than style. In particular, \texttt{res4} is followed by an average-pooling layer, which essentially forwards the mean vector to the prediction layer so it captures label-related information. As a consequence, mixing the statistics at \texttt{res4} breaks the inherent label space. This is clearer in Fig.~\ref{fig:tsne_feat_sty}: the features and style statistics in \texttt{res1-3} exhibit clustering patterns based on domains while those in \texttt{res4} have a high correlation with class labels.

\begin{table}[t]
    \tabstyle{5pt}
    \caption{Examining MixStyle's design choices.}
    \subfloat[]{
    \label{tab:mix_vs_replace}
        \begin{tabular}{lc}
        \toprule
        Design & Accuracy \\
        \midrule
        Mixing & \textbf{82.8}$\pm$0.4 \\
        Replacing & 82.1$\pm$0.5 \\
        \bottomrule
        \end{tabular}
    }
    \quad \quad
    \subfloat[]{
    \label{tab:rand_vs_fixed_shuffle}
        \begin{tabular}{lc}
        \toprule
        Design & Accuracy \\
        \midrule
        Random & \textbf{82.8}$\pm$0.4 \\
        Fixed & 82.4$\pm$0.5 \\
        \bottomrule
        \end{tabular}
    }
\end{table}

\begin{table*}[t]
    \tabstyle{5pt}
    \caption{Style mixing between random samples vs.~samples of the same class.}
    \label{tab:style_mix_same_class}
    \begin{tabular}{l ccccc}
    \toprule
    & Art painting & Cartoon & Photo & Sketch & Avg. \\
    \midrule
    ERM & 77.0 & 75.9 & 96.0 & 69.2 & 79.5 \\
    MixStyle (random) & 82.3 & 79.0 & 96.3 & 73.8 & 82.8 \\
    MixStyle (same-class) & 79.3 & 76.3 & 94.7 & 68.5 & 79.7 \\
    \bottomrule
    \end{tabular}
\end{table*}

\minisec{Mixing vs.~Replacing}
Unlike the AdaIN formulation that completely replaces one style with another, MixStyle gives a more general formulation that mixes two styles with a random convex combination. Table~\ref{tab:mix_vs_replace} suggests that mixing is a better choice---mixing produces more diverse styles (imagine an interpolation between two distant datapoints).

\minisec{Random vs.~Fixed Shuffle}
Table~\ref{tab:where_mixstyle} confirms that applying MixStyle to multiple layers is better, but this may raise another question over whether to shuffle the mini-batch at different layers or just follow the same shuffled order at all selected layers. Table~\ref{tab:rand_vs_fixed_shuffle} advocates the random shuffle design, probably due to the increased noise level that enhances regularization.

\minisec{Random vs.~Same-class Style Mixing}
We conduct an experiment to evaluate which style mixing strategy is better: style mixing among random samples vs.~style mixing within the same class. The experiment is done on PACS and the results are shown on Table~\ref{tab:style_mix_same_class}. It is clear that the ``same-class'' version's performance is similar to the ERM model, which suggests that style mixing between samples of the same class is ineffective.

\begin{figure*}[t]
    \centering
    \includegraphics[width=.9\textwidth]{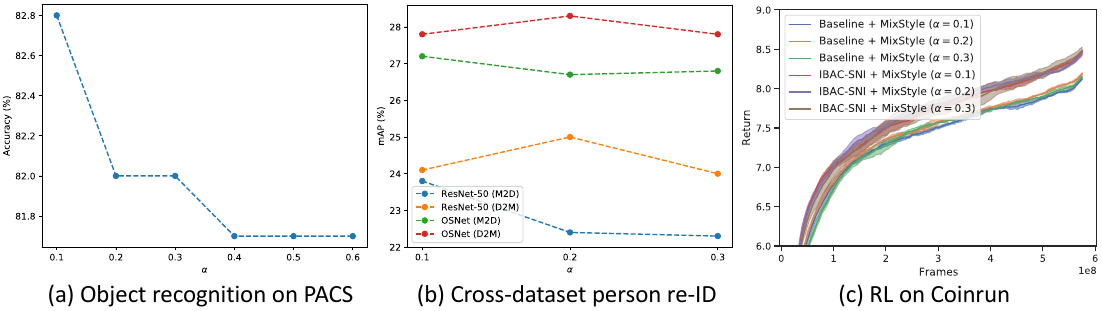}
    \caption{Evaluating $\alpha$, the hyper-parameter of the beta distribution (best viewed by zoom-in). M and D in (b) denote Market1501 and Duke, respectively.}
    \label{fig:ablation_alpha}
\end{figure*}

\minisec{Beta Distribution's Parameter}
In addition to positions to insert MixStyle, another hyper-parameter is $\alpha$, the beta distribution's shape parameter that controls the distribution of the convex weight $\lambda$ in Eq.~\eqref{eq:mixstyle}. A smaller $\alpha$ means $\lambda$ is biased toward the extreme value of 0 or 1, while a larger $\alpha$ means $\lambda$ is more likely to sit around 0.5. The study is summarized in Fig.~\ref{fig:ablation_alpha}. On PACS, the accuracy slides from 82.8\% to 81.7\% with $\alpha$ increasing from 0.1 to 0.4, but stabilizes thereafter. The results suggest that the performance is not too sensitive to $\alpha$, and choosing $\alpha \in \{0.1, 0.2, 0.3\}$ seems to be a good start point. The results on the re-ID and RL task further confirm that the variance for different $\alpha$'s within $\{0.1, 0.2, 0.3\}$ is generally small. Overall, $\alpha = 0.1$ is a good default setting.

\begin{figure}[t]
    \centering
    \includegraphics[width=\columnwidth]{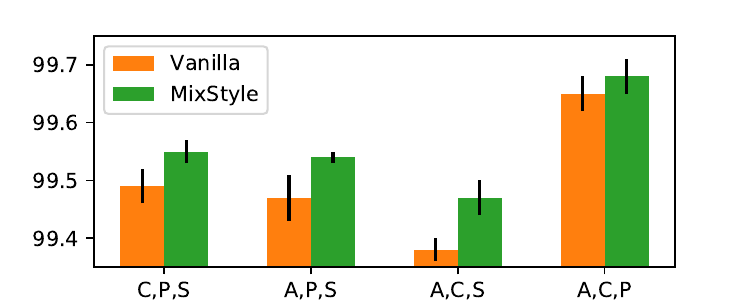}
    \caption{Test results on the source domains on PACS.}
    \label{fig:srcdom_perform}
\end{figure}

\minisec{Source Domain Performance}
Fig.~\ref{fig:srcdom_perform} justifies that MixStyle does not sacrifice the performance on seen domains in exchange for gains on unseen domains. In fact, the recognition accuracy on the source domains is improved but very slightly (the results are saturated). Such non-trivial gains on source domains well explain the reason why MixStyle works in UDA (Section~\ref{sec:exp;ssec:uda})---the unlabeled target domain could be viewed as a special ``source'' since its images are seen during training.

\section{Discussion and Conclusion}

MixStyle bypasses image synthesis in the pixel space by exploiting the close relation between visual domains and feature statistics to achieve highly-efficient data augmentation. The non-trivial design based on simply mixing feature statistics between instances yields significant improvements to OOD generalization performance---even when only a few labels are available. With a simple modification to the mixing strategy, MixStyle shows potential in dealing with unlabeled data, such as the scenarios of semi-supervised domain generalization and unsupervised domain adaptation.

To unleash the power of MixStyle, we provide practical guidelines on how to integrate MixStyle into a CNN system, e.g., one should apply MixStyle to multiple shallow CNN layers. However, for a new task, it remains relatively unclear exactly to which layers MixStyle should be applied. Nonetheless, this is arguably the only ``hyper-parameter'' that needs to be ``tuned'', which is far less than most existing domain generalization methods. To completely eliminate the manual selection process, one could explore, for example, applying MixStyle to every plausible layer and using a differentiable architecture search algorithm to automatically identify the optimal set of layers to choose~\cite{liu2019darts,zhou2021osnet}.

From the application point of view, MixStyle shows encouraging results on a wide variety of problems including object recognition, person re-identification, and reinforcement learning. The extensive experiments conducted in this work suggest that MixStyle can cope with domain shifts related to colors, textures, illuminations, backgrounds, and other visual factors that are associated with image style. This makes sense because based on what we have learned from the image style transfer literature~\cite{huang2017arbitrary,dumoulin2017learned}, mixing feature statistics can well simulate these visual changes and thus help to learn representations invariant to them. It is worth mentioning that MixStyle has already been used by the community in a variety of AI applications e.g., vehicle re-identification~\cite{huynh2021strong}, semantic segmentation~\cite{zhao2021source}, 3D point clouds processing~\cite{zhao2022crossmodal}, medical image analysis~\cite{kushibarcancer}, speech recognition~\cite{schmid2022cp}, and so on.

In terms of shortcomings, MixStyle might be less effective in dealing with geometrical shifts, such as rotation or changing viewpoints. This is probably why MixStyle's gains on Office-Home are less appealing---Fig.~\ref{fig:im_pacs_oh_reid}(b) shows that the domain shifts on Office-Home are more concerned with viewpoint rather than image style (the bed scenes in the domain of art, product and real-world look similar except viewpoints).

\section*{Data Availability Statement}
The datasets generated during and/or analysed during the current study are available in the project's github repository, \url{https://github.com/KaiyangZhou/mixstyle-release}.


%
%

\bibliographystyle{spmpsci}      
\bibliography{ref}   

\end{document}